\documentclass[11pt,a4paper]{article}
\usepackage[hyperref]{acl2020}
\usepackage{times}
\usepackage{stackengine}
\usepackage{hyperref}
\RequirePackage{latexsym,amsmath,amsthm,amssymb,bm}
\RequirePackage{enumitem}
\setlist[itemize]{leftmargin=*,labelwidth=0pt,nosep}
\setlist[enumerate]{nosep}
\setlist[description]{nosep}
\RequirePackage[textsize=scriptsize]{todonotes}
\usepackage{microtype}
\usepackage{caption}

\makeatletter
\g@addto@macro{\normalsize}{%
\setlength{\abovedisplayskip}{0pt}%
\setlength{\abovedisplayshortskip}{0pt}%
\setlength{\belowdisplayskip}{0pt}%
\setlength{\belowdisplayshortskip}{0pt}}
\makeatother

\newcommand{\shortname}{{\scshape IMoJIE}}
\newcommand{\longname}{Iterative Memory-Based Joint Open Information Extraction}
\newcommand{\boldlongname}{\textbf{I}terative \textbf{M}em\textbf{O}ry \textbf{J}oint Open \textbf{I}nformation \textbf{E}xtraction}
\newcommand{\boldshortname}{{\sc \textbf{IMoJIE}}}

\aclfinalcopy 
\def\aclpaperid{1616} 

\def\intitle{\shortname: \longname}

\title{\intitle}

\author{
  Keshav Kolluru\textsuperscript{1},  Samarth Aggarwal\textsuperscript{1}, Vipul Rathore\textsuperscript{1}, Mausam\textsuperscript{1} and Soumen Chakrabarti\textsuperscript{2}\\
  \textsuperscript{1} Indian Institute of Technology Delhi\\
  \texttt{\{keshav.kolluru, samarth.aggarwal.2510, rathorevipul28\}@gmail.com} \\
  \texttt{mausam@cse.iitd.ac.in} \\
  \textsuperscript{2} Indian Institute of Technology Bombay\\
  \texttt{soumen@cse.iitb.ac.in }
}
\date{}

\begin{document}
\maketitle
\begin{abstract}
While traditional systems for Open Information Extraction were statistical and rule-based, recently neural models have been introduced for the task. Our work builds upon CopyAttention, a sequence generation OpenIE model \cite{cui&al18}. Our analysis reveals that CopyAttention produces a constant number of extractions per sentence, and its extracted tuples often express redundant information.

We present \shortname, an extension to CopyAttention, which produces the next extraction conditioned on all previously extracted tuples. This approach overcomes both shortcomings of CopyAttention, resulting in a variable number of diverse extractions per sentence. We train \shortname{} on training data bootstrapped from extractions of several non-neural systems, which have been automatically filtered to reduce redundancy and noise.  \shortname{} outperforms CopyAttention by about 18 F1 pts, and a BERT-based strong baseline by 2 F1 pts, establishing a new state of the art for the task. 

\end{abstract}

\begin{table*}
\centering
{\small
\begin{tabular}{|c|l|}
\hline
\textbf{Sentence} & \begin{tabular}[c]{@{}l@{}}He was appointed Commander of the Order of the British Empire in the 1948\\  Queen's Birthday Honours and was knighted in the 1953 Coronation Honours .\end{tabular} \\ \hline
\textbf{CopyAttention} & \begin{tabular}[c]{@{}l@{}}( He ; was appointed ; Commander ... Birthday Honours )\\( He ; was appointed ; Commander ... Birthday Honours and was knighted ... Honours )\\( Queen 's Birthday Honours ; was knighted ; in the 1953 Coronation Honours )\\( He ; was appointed ; Commander of the Order of the British Empire in the 1948 )\\( the 1948 ; was knighted ; in the 1953 Coronation Honours)\end{tabular} \\ \hline
\textbf{\begin{tabular}[c]{@{}l@{}}IMOJIE\\ \end{tabular}} & \begin{tabular}[c]{@{}l@{}}( He ; was appointed ; Commander of the Order ... Birthday Honours )\\( He ; was knighted ; in the 1953 Coronation Honours )\end{tabular} \\ \hline
\end{tabular}
}
\caption{\shortname~ vs. CopyAttention. CopyAttention suffers from stuttering, which \shortname{} does not.}
\label{tab:redundancy-example}
\end{table*}

\begin{table*}
\centering
{\small
\begin{tabular}{|c|l|}
\hline
\textbf{Sentence} & \begin{tabular}[c]{@{}l@{}}Greek and Roman pagans , who saw their relations with the gods in political and social\\  terms , scorned the man who constantly trembled with fear at the thought of the gods ,\\  as a slave might fear a cruel and capricious master .\end{tabular} \\ \hline
\textbf{OpenIE-4} & ( the man ;  constantly trembled ;   ) \\ \hline
\textbf{\begin{tabular}[c]{@{}c@{}}IMOJIE\\ \end{tabular}} & \begin{tabular}[c]{@{}l@{}}( a slave ; might fear ; a cruel and capricious master )\\ ( Greek and Roman pagans ; scorned ; the man who ... capricious master )\\ ( the man ; constantly trembled ; with fear at the thought of the gods )\\ ( Greek and Roman pagans ; saw ; their relations with the gods in political and social terms )\end{tabular} \\ \hline
\end{tabular}
}
\caption{\shortname~ vs. OpenIE-4. Pipeline nature of OpenIE-4 can get confused by long convoluted sentences, but \shortname{} responds gracefully.}
\label{tab:quality-example}
\end{table*}

\section{Introduction}

Extracting structured information from unstructured text has been a key research area within NLP. The paradigm of Open Information Extraction (OpenIE) \cite{banko&al07} uses an open vocabulary to convert natural text to semi-structured representations, by extracting a set of (subject, relation, object) tuples. OpenIE has found wide use in many downstream NLP tasks \cite{mausam16} like multi-document question answering and summarization \cite{fan&al19}, event schema induction \cite{balasubramanian-emnlp13} and word embedding generation \cite{stanovsky&al15}.

Traditional OpenIE systems are statistical or rule-based. They are largely unsupervised in nature, or bootstrapped from extractions made by earlier systems. They often consist of several components like POS tagging, and syntactic parsing.  To bypass error accumulation in such pipelines, end-to-end neural systems have been proposed recently.

Recent neural OpenIE methods belong to two categories: sequence \emph{labeling}, e.g., RnnOIE \citep{stanovsky&al18} and sequence \emph{generation}, e.g., CopyAttention \citep{cui&al18}.  In principle, generation is more powerful because it can introduce auxiliary words or change word order.  However, our analysis of CopyAttention reveals that it suffers from two drawbacks.  First, it does not naturally adapt the number of extractions to the length or complexity of the input sentence.  Second, it is susceptible to \emph{stuttering}: extraction of multiple triples bearing redundant information.

These limitations arise because its decoder has no explicit mechanism to remember what parts of the sentence have already been `consumed' or what triples have already been generated. Its decoder  uses a fixed-size beam for inference. However, beam search can only ensure that the extractions are not exact duplicates.


In response, we design the first neural OpenIE system that uses sequential decoding of tuples conditioned on previous tuples. We achieve this by adding every generated extraction so far to the encoder. This iterative process stops when the \textit{EndOfExtractions} tag is generated by the decoder, allowing it to produce a variable number of extractions. We name our system \boldlongname\ (\boldshortname).

CopyAttention uses a bootstrapping strategy, where the extractions from  OpenIE-4 \cite{christensen&al11,pal&al16} are used as training data. However, we believe that training on extractions of multiple systems is preferable. For example, OpenIE-4 benefits from high precision compared to ClausIE \cite{corro&al13}, which offers high recall. By aggregating extractions from both, \shortname{} could potentially obtain a better precision-recall balance.

However, simply concatenating extractions from multiple systems does not work well, as it leads to redundancy as well as exaggerated noise in the dataset. 
We devise an unsupervised \textbf{Score-and-Filter} mechanism to automatically select a subset of these extractions that are non-redundant and expected to be of high quality. Our approach scores all extractions with a scoring model, followed by filtering to reduce redundancy.

\begin{figure*}[t]
\includegraphics[width=.99\hsize]{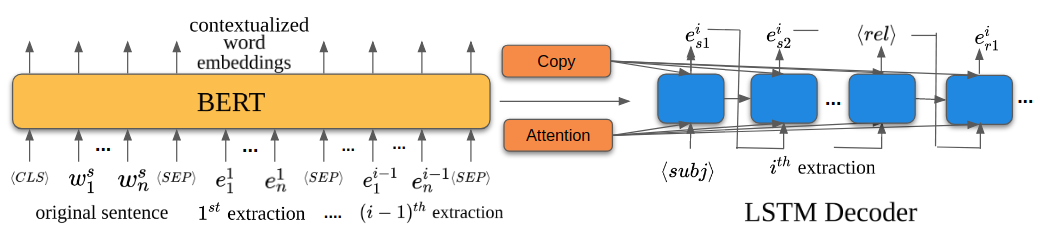}
\caption{One step of the sequential decoding process, for generating the $i^\text{th}$ extraction, which takes the original sentence and all extractions numbered $1,\ldots,i-1$, previously generated, as input.}
\label{fig:seq_dec}
\end{figure*}

We compare \shortname\ against several neural and non-neural systems, including our extension of CopyAttention that uses BERT \cite{devlin&al19} instead of an LSTM at encoding time, which forms a very strong baseline. On the recently proposed CaRB metric, which penalizes redundant extractions \cite{bhardwaj&al19}, \shortname\ outperforms CopyAttention by about 18 pts in F1 and our strong BERT baseline by 2 pts, establishing a new state of the art for OpenIE. We release \shortname\ \& all related resources for further research\footnote{\protect\url{https://github.com/dair-iitd/imojie}}.  In summary, our contributions are:
\begin{itemize}
\item We propose \shortname{}, a neural OpenIE system that generates the next extraction, fully conditioned on the extractions produced so far. \shortname{} produce a variable number of diverse extractions for a sentence,
\item We present an unsupervised aggregation scheme to bootstrap training data by combining extractions from multiple OpenIE systems.
\item \shortname{} trained on this data establishes a new SoTA in OpenIE, beating previous systems and also our strong BERT-baseline.
\end{itemize}

\section{Related Work}

Open Information Extraction (OpenIE) involves extracting (arg1 phrase, relation phrase, arg2 phrase) assertions from a sentence. Traditional open extractors are rule-based or statistical, e.g., Textrunner \cite{banko&al07}, ReVerb \cite{Fader&al11,etzioni-ijcai11}, OLLIE \cite{Mausam&al12}, Stanford-IE \cite{angeli&al15}, ClausIE \cite{corro&al13}, OpenIE-4 \cite{christensen&al11,pal&al16}, OpenIE-5 \cite{Saha&al2017, Saha2018OpenIE}, PropS \cite{Stanovsky&al2016}, and MinIE \cite{gashteovski&al17}. These use syntactic or semantic parsers combined with rules to extract tuples from sentences. 

Recently, to reduce error accumulation in these pipeline systems, neural OpenIE models have been proposed. They belong to one of two paradigms: sequence \textit{labeling} or sequence \textit{generation}. 

\vspace{0.5ex}
\noindent \textbf{Sequence Labeling} involves tagging each word in the input sentence as belonging to the subject, predicate, object or other. The final extraction is obtained by collecting labeled spans into different fields and constructing a tuple.  RnnOIE \cite{stanovsky&al18} is a labeling system that first identifies the relation words and then uses sequence labelling to get their arguments. It is trained on OIE2016 dataset, which postprocesses SRL data for OpenIE \cite{Stanovsky2016EMNLP}.

SenseOIE \cite{roy&al19}, improves upon RnnOIE by using the extractions of multiple OpenIE systems as features in a sequence labeling setting. However, their training requires manually annotated gold extractions, which is not scalable for the task. This restricts SenseOIE to train on a dataset of 3,000 sentences. In contrast, our proposed \textit{Score-and-Filter} mechanism is unsupervised and can scale unboundedly. \citet{jiang12019improving} is another labeling system that better calibrates extractions across sentences.

SpanOIE \cite{zhan&al19} uses a span selection model, a variant of the sequence labelling paradigm. Firstly, the predicate module finds the predicate spans in a sentence. Subsequently, the argument module outputs the arguments for this predicate. However, SpanOIE cannot extract nominal relations. Moreover, it bootstraps its training data over a single OpenIE system only. In contrast, \shortname{} overcomes both of these limitations.

\vspace{0.5ex}
\noindent \textbf{Sequence Generation} uses a Seq2Seq model to generate output extractions one word at a time. The generated sequence contains field demarcators, which are used to convert the generated flat sequence to a tuple.
CopyAttention \cite{cui&al18} is a neural generator trained over bootstrapped data generated from OpenIE-4 extractions on a large corpus. During inference, it uses beam search to get the predicted extractions. It uses a fixed-size beam, limiting it to output a constant number of extractions per sentence. Moreover, our analysis shows that CopyAttention extractions severely lack in diversity, as illustrated in Table~\ref{tab:redundancy-example}.

\citet{sun2018logician} propose the \emph{Logician} model, a restricted sequence generation model for extracting tuples from Chinese text. Logician relies on coverage attention and gated-dependency attention, a language-specific heuristic for Chinese. Using coverage attention, the model also tackles generation of multiple extractions while being globally-aware. We compare against Logician's coverage attention as one of the approaches for increasing diversity. 


Sequence-labeling based models lack the ability to change the sentence structure or introduce new auxiliary words while uttering predictions. For example, they cannot extract (Trump, is the President of, US) from ``US President Trump", since `is', `of' are not in the original sentence. 
On the other hand, sequence-generation models are more general and, in principle, need not suffer from these limitations.


\paragraph{Evaluation:} All neural models have shown improvements over the traditional systems using the OIE2016 benchmark. However, recent work shows that the OIE2016 dataset is quite noisy, and that its evaluation does not penalize highly redundant extractions \cite{william&al18}. 
In our work, we use the latest CaRB benchmark, which crowdsources a new evaluation dataset, and also provides a modified evaluation framework to downscore near-redundant extractions \cite{bhardwaj&al19}.

\begin{figure*}[t]
\includegraphics[width=.99\hsize]{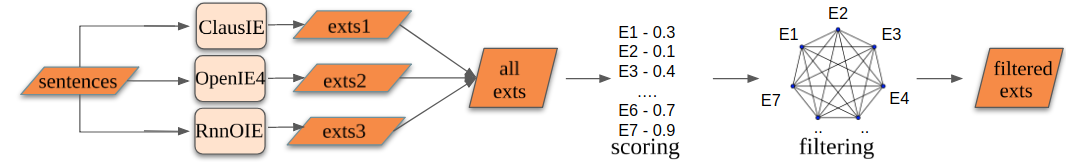}
\caption{Ranking-Filtering subsystem for combining extractions from multiple open IE systems in an unsupervised fashion.  (`Exts'=extractions.)}
\label{fig:score_filter}
\end{figure*}

\section{Sequential Decoding}


We now describe \shortname{}, our generative approach that can output a variable number of diverse extractions per sentence. The architecture of our model is illustrated in Figure \ref{fig:seq_dec}. At a high level, the next extraction from a sentence is best determined in context of all other tuples extracted from it so far.  Hence, \shortname{} uses a decoding strategy that generates extractions in a sequential fashion, one after another, each one being aware of all the ones generated prior to it.




This kind of sequential decoding is made possible by the use of an \textit{iterative memory}. Each of the generated extractions are added to the memory so that the next iteration of decoding has access to all of the previous extractions. We simulate this iterative memory with the help of BERT encoder, whose input includes the [\textit{CLS}] token and original sentence appended with the decoded extractions so far, punctuated by the separator token [\textit{SEP}] before each extraction.


\shortname{} uses an LSTM  decoder, which is initialized with the embedding of [\textit{CLS}] token. The contextualized-embeddings of all the word tokens are used for the Copy \citep{gu&al16} and Attention \citep{bahdanau&al15} modules. The decoder generates the tuple one word at a time, producing $\langle \textit{rel} \rangle$ and $\langle \textit{obj} \rangle$ tokens to indicate the start of relation and object respectively. The iterative process continues until the \textit{EndOfExtractions} token is generated.

The overall process can be summarized as:
\begin{enumerate}
    \item Pass the sentence through the Seq2Seq architecture to generate the first extraction.
    \item Concatenate the generated extraction with the existing input and pass it again through the Seq2Seq architecture to generate the next extraction.
    \item Repeat Step 2 until the \textit{EndOfExtractions} token is generated.
\end{enumerate}

\shortname{} is trained using a cross-entropy loss between the generated output and the gold output.

\section{Aggregating Bootstrapped  Data}
\label{Aggregating Bootstrapping data}

\subsection{Single Bootstrapping System}

To train generative neural models for the task of OpenIE, we need a set of sentence-extraction pairs. It is ideal to curate such a training dataset via human annotation, but that is impractical, considering the scale of training data required for a neural model. We follow \citet{cui&al18}, and use  bootstrapping --- using extractions from a pre-existing OpenIE system as `silver'-labeled (as distinct from `gold'-labeled) instances to train the neural model. We first order all extractions in the decreasing order of confidences output by the original system. We then construct training data in \shortname's input-output format, assuming that this is the order in which it should produce its extractions.

\subsection{Multiple Bootstrapping Systems}
Different OpenIE systems have diverse quality characteristics. For example, the human-estimated (precision, recall) of OpenIE-4 is $(\textbf{61},43)$ while that of ClausIE is $(40,\textbf{50})$. Thus, by using their combined extractions as the bootstrapping dataset, we might potentially benefit from the high precision of OpenIE-4 and high recall of ClausIE.

However, simply pooling all extractions would not work, because of the following serious hurdles.
\begin{description}
\item[No calibration:] Confidence scores assigned by different systems are not calibrated to a comparable scale.
\item[Redundant extractions:] Beyond exact duplicates, multiple systems produce similar extractions with low marginal utility.
\item[Wrong extractions:] Pooling inevitably pollutes the silver data and can amplify incorrect instances, forcing the downstream open IE system to learn poor-quality extractions.
\end{description}
We solve these problems using a \textbf{Score-and-Filter} framework, shown in Figure~\ref{fig:score_filter}.

\vspace{0.5ex}
\noindent \textbf{Scoring:}
All systems are applied on a given sentence, and the pooled set of extractions are scored such that good (correct, informative) extractions generally achieve higher values compared to bad (incorrect) and redundant ones.  In principle, this score may be estimated by the generation score from \shortname, trained on a single system. 
In practice, such a system is likely to consider extractions similar to its bootstrapping training data as good, while disregarding extractions of other systems, even though those extractions may also be of high quality. 
To mitigate this bias, we use an \shortname{} model, pre-trained on a \textit{random bootstrapping dataset}. The random bootstrapping dataset is generated by picking extractions for each sentence randomly from any one of the bootstrapping systems being aggregated. We assign a score to each extraction in the pool based on the confidence value given to it by this \shortname~(Random) model.

\vspace{0.5ex}
\noindent \textbf{Filtering:}
We now filter this set of extractions for redundancy. 
Given the set of ranked extractions in the pool, we wish to select that subset of extractions that have the best confidence scores (assigned by the random-boostrap model), while having minimum similarity to the other selected extractions.

We model this goal as the selection of an optimal subgraph from a suitably designed complete weighted graph. Each node in the graph corresponds to one extraction in the pool. Every pair of nodes $(u,v)$ are connected by an edge. Every edge has an associated weight $R(u,v)$ signifying the similarity between the two corresponding extractions. Each node $u$ is assigned a score $f(u)$ equal to the confidence given by the random-bootstrap model.

Given this graph $G = (V, E)$ of all pooled extractions of a sentence, we aim at selecting a subgraph $G' = (V', E')$ with $V' \subseteq V$, such that the most significant ones are selected, whereas the extractions redundant with respect to already-selected ones are discarded. Our objective is
\begin{equation} \label{eq:1}
\max_{G'\subseteq G} \sum_{i=1}^{|V'|} f(u_i) -
\sum_{j=1}^{|V'|-1}\sum_{k=j+1}^{|V'|} R(u_{j},u_{k}),
\end{equation}
where $u_{i}$ represents node $i\in V'$.  
We compute $R(u,v)$ as the ROUGE2 score between the serialized triples represented by nodes $u$ and $v$. We can intuitively understand the first term as the aggregated sum of significance of all selected triples and second term as the redundancy among these triples.

If $G$ has $n$ nodes, we can pose the above objective as:
\begin{align}
\max_{\bm{x} \in \{0,1\}^{n}}  \bm{x}^\top \bm{f} - \bm{x}^\top \bm{R} \bm{x},
\end{align}
where $\bm{f} \in \mathbb{R}^n$ representing the node scores, i.e., $f[i]=f(u_{i})$, and
$\bm{R} \in \mathbb{R}^{n\times n}$ is a symmetric matrix with entries $R_{j,k}=\text{ROUGE2}(u_{j},u_{k})$.
$\bm{x}$ is the decision vector, with $x[i]$ indicating whether a particular node $u_i \in V'$ or not. This is an instance of Quadratic Boolean Programming and is NP-hard, but in our application $n$ is modest enough that this is not a concern.  We use the QPBO (Quadratic Pseudo Boolean Optimizer) solver\footnote{\href{https://pypi.org/project/thinqpbo/}{https://pypi.org/project/thinqpbo/}} \cite{Rother2007OptimizingBM} to find the optimal $\bm{x}^*$ and recover~$V'$.

\section{Experimental Setup}
\label{sec:ExptSetup}

\subsection{Training Data Construction}

We obtain our training sentences by scraping Wikipedia, because Wikipedia is a comprehensive source of informative text from diverse domains, rich in entities and relations.  
Using sentences from Wikipedia ensures that our model is not biased towards data from any single domain.

We run OpenIE-4\footnote{\href{https://github.com/knowitall/openie}{https://github.com/knowitall/openie}}, ClausIE\footnote{\href{https://www.mpi-inf.mpg.de/departments/databases-and-information-systems/software/clausie}{https://www.mpi-inf.mpg.de/clausie}} and RnnOIE\footnote{\href{https://github.com/gabrielStanovsky/supervised-oie}{https://github.com/gabrielStanovsky/supervised-oie}} on these sentences to generate a set of OpenIE tuples for every sentence, which are then ranked and filtered using our Score-and-Filter technique. 
These tuples are further processed to generate training instances in \shortname's input-output format. 

Each sentence contributes to multiple (input, output) pairs for the \shortname{} model. The first training instance contains the sentence itself as input and the first tuple as output. For example, (``I ate an apple and an orange.'', ``I; ate; an apple''). The next training instance, contains the sentence concatenated with previous tuple as input and the next tuple as output (``I ate an apple and an orange. [SEP] I; ate; an apple'',  ``I; ate; an orange''). The final training instance generated from this sentence includes all the extractions appended to the sentence as input and \textit{EndOfExtractions} token as the output. Every sentence gives the seq2seq learner one training instance more than the number of tuples. 

While forming these training instances, the tuples are considered in decreasing order of their confidence scores.  If some OpenIE system does not provide confidence scores for extracted tuples, then the output order of the tuples may be used.  

\subsection{Dataset and Evaluation Metrics}

We use the CaRB data and evaluation framework \cite{bhardwaj&al19} to evaluate the systems\footnote{Our reported CaRB scores for OpenIE-4 and OpenIE-5 are slightly different from those reported by \citet{bhardwaj&al19}. The authors of CaRB have verified our values.} at different confidence thresholds, yielding a precision-recall curve. We identify three important summary metrics from the P-R curve. 

\noindent \textbf{Optimal F1}: We find the point in the P-R curve corresponding to the largest F1 value and report that.  This is the operating point for getting extractions with the best precision-recall trade-off.

\noindent \textbf{AUC}: This is the area under the P-R curve. This metric is useful when the downstream application can use the confidence value of the extraction.

\noindent \textbf{Last F1}: This is the F1 score computed at the point of zero confidence. This is of importance when we cannot compute the optimal threshold, due to lack of any gold-extractions for the domain. Many downstream applications of OpenIE, such as text comprehension \citep{stanovsky&al15} and sentence similarity estimation \citep{janara&al14}, use \emph{all} the extractions output by the OpenIE system. Last~F1 is an important measure for such applications.

\subsection{Comparison Systems}
We compare \shortname{} against several non-neural baselines, including Stanford-IE, OpenIE-4, OpenIE-5, ClausIE, PropS, MinIE, and OLLIE. We also compare against the sequence labeling baselines of RnnOIE, SenseOIE, and the span selection baseline of SpanOIE. Probably the most closely related baseline to us is the neural generation baseline of CopyAttention. To increase CopyAttention's diversity, we compare against an English version of Logician, which adds coverage attention to a single-decoder model that emits all extractions one after another. We also compare against CopyAttention augmented with diverse beam search \cite{diversebeam} --- it adds a diversity term to the loss function so that new beams have smaller redundancy with respect to all previous beams.

Finally, because our model is based on BERT, we reimplement CopyAttention with a BERT encoder --- this forms a very strong baseline for our task. 

\begin{table}[t]
\centering {\footnotesize
\begin{tabular}{lccc}
\hline
\textbf{System} & \multicolumn{3}{c}{\textbf{Metric}}  \\
 & \multicolumn{1}{c}{Opt. F1} & \multicolumn{1}{c}{AUC} & \multicolumn{1}{c}{Last F1}\\
\hline
Stanford-IE & 23 & 13.4 & 22.9 \\
OllIE & 41.1 & 22.5 & 40.9 \\ 
PropS & 31.9 & 12.6 & 31.8 \\ 
MinIE & 41.9 & -$^*$ & 41.9 \\ 
OpenIE-4 & 51.6 & 29.5 & 51.5 \\ 
OpenIE-5 & 48.5 & 25.7 & 48.5 \\ 
ClausIE & 45.1 & 22.4 & 45.1 \\ \hline 
CopyAttention & 35.4 & 20.4 & 32.8 \\ 
RNN-OIE & 49.2 & 26.5 & 49.2 \\ 
Sense-OIE & 17.2 & -$^*$ & 17.2 \\ 
Span-OIE & 47.9 & -$^*$ & 47.9 \\
CopyAttention + BERT & 51.6 & 32.8 & 49.6 \\ \hline
\shortname~ & \textbf{53.5} & \textbf{33.3} & \textbf{53.3} \\ \hline
\end{tabular} }
\caption{Comparison of various OpenIE systems - non-neural, neural and proposed models. (*)~Cannot compute AUC as Sense-OIE, MinIE do not emit confidence values for extractions and released code for Span-OIE does not provision calculation of confidence values. In these cases, we report the Last F1 as the Opt. F1}
\label{tab:scores-table}
\end{table}

\begin{table}[t]
\centering {\footnotesize
\begin{tabular}{lccc}
\hline
\textbf{System} & \multicolumn{3}{c}{\textbf{Metric}}  \\
 & \multicolumn{1}{c}{Opt. F1} & \multicolumn{1}{c}{AUC} & \multicolumn{1}{c}{Last F1}\\
\hline
CopyAttention & 35.4 & 20.4 & 32.8 \\ 
CoverageAttention & 41.8 & 22.1 & 41.8 \\ 
CoverageAttention+BERT & 47.9 & 27.9 & 47.9 \\ 
Diverse Beam Search & 46.1 & 26.1 & 39.6 \\ 
\shortname~ (w/o BERT) & 37.9 & 19.1 & 36.6 \\ 
\shortname~ & \textbf{53.2} & \textbf{33.1} & \textbf{52.4} \\ \hline
\end{tabular} }
\caption{Models to solve the redundancy issue prevalent in Generative Neural OpenIE systems. All systems are bootstrapped on OpenIE-4.}
\label{tab:redundancy_rem_exp}
\end{table}

\begin{table}[t]
\centering {\footnotesize
\begin{tabular}{lccc}
\hline
\textbf{Bootstrapping } & \multicolumn{3}{c}{\textbf{Metric}}  \\
\textbf{Systems} & \multicolumn{1}{c}{Opt. F1} & \multicolumn{1}{c}{AUC} & \multicolumn{1}{c}{Last F1} \\
\hline
ClausIE & 49.2 & 31.4 & 45.5 \\
RnnOIE & 51.3 & 31.1 & 50.8 \\
OpenIE-4 & 53.2 & 33.1 & 52.4 \\
OpenIE-4+ClausIE & 51.5 & 32.5 & 47.1 \\ 
OpenIE-4+RnnOIE & 53.1 & 32.1 & 53.0 \\ 
ClausIE+RnnOIE & 50.9 & 32.2 & 49.8 \\ 
All & \textbf{53.5} & \textbf{33.3} & \textbf{53.3} \\ \hline
\end{tabular} }
\caption{\shortname{} trained with different combinations of bootstrapping data from 3 systems - OpenIE-4, ClausIE, RNNOIE. Graph filtering is not used over single datasets.}
\label{tab:dataset-ablation}
\end{table}

\subsection{Implementation}

We implement \shortname{} in the AllenNLP framework\footnote{\href{https://github.com/allenai/allennlp}{https://github.com/allenai/allennlp}} \citep{gardner2018allennlp} using Pytorch~1.2.  We use ``BERT-small'' model for faster training. Other hyper-parameters include learning rate for BERT, set to $2\times10^{-5}$, and learning rate, hidden dimension, and word embedding dimension of the decoder LSTM, set to $(10^{-3}, 256, 100)$, respectively.

Since the model or code of CopyAttention \citep{cui&al18} were not available, we implemented it ourselves. Our implementation closely matches their reported scores, achieving (F1, AUC) of (56.4, 47.7) on the OIE2016 benchmark.


\section{Results and Analysis}
\label{sec:ExptResult}
\vspace{0.5ex}
\subsection{Performance of Existing Systems}
\emph{How well do the neural systems perform as compared to the rule-based systems?}

Using CaRB evaluation, we find that, contrary to previous papers, neural OpenIE systems are not necessarily better than prior non-neural systems (Table~\ref{tab:scores-table}).  Among the systems under consideration, the best non-neural system reached Last~F1 of 51.5, whereas the best existing neural model could only reach~49.2. Deeper analysis reveals that CopyAttention produces redundant extractions conveying nearly the same information, which CaRB effectively penalizes.  RnnOIE performs much better, however suffers due to its lack of generating auxilliary verbs and implied prepositions. Example, it can only generate (Trump; President; US) instead of (Trump; is President of; US) from the sentence ``US President Trump...". Moreover, it is trained only on limited number of pseudo-gold extractions, generated by \citet{michael&al18}, which does not take advantage of boostrapping techniques.

\begin{table}
\centering {\footnotesize
\begin{tabular}{lccc}
\hline
 \textbf{Filtering} & \multicolumn{3}{c}{\textbf{Metric}}  \\
 & \multicolumn{1}{c}{Opt. F1} & \multicolumn{1}{c}{AUC} & \multicolumn{1}{c}{Last F1}\\
\hline
None & 49.7 & \textbf{34.5} & 37.4 \\ 
Extraction-based & 46 & 29.2 & 44.9 \\ 
Sentence-based &  49.5 & 32.7 & 48.6\\ 
Score-And-Filter & \textbf{53.5} & 33.3 & \textbf{53.3} \\ \hline
\end{tabular} }
\caption{Performance of \shortname{} on aggregated dataset \textbf{OpenIE-4+ClausIE+RnnOIE}, with different filtering techniques. For comparison, SenseOIE trained on multiple system extractions gives an F1 of 17.2 on CaRB.}
\label{tab:dataset-aggregation1}
\end{table}

\vspace{0.5ex}
\subsection{Performance of \shortname}
\emph{How does \shortname{} perform compared to the previous neural and rule-based systems?} 

In comparison with existing neural and non-neural systems, \shortname{} trained on aggregated bootstrapped data performs the best.  It outperforms OpenIE-4, the best existing OpenIE system, by 1.9 F1 pts, 3.8 pts of AUC, and 1.8 pts of Last-F1. Qualitatively, we find that it makes fewer mistakes than OpenIE-4, probably because OpenIE-4 accumulates errors from upstream parsing modules (see Table \ref{tab:quality-example}). 

\shortname{} outperforms CopyAttention by large margins -- about 18 Optimal F1 pts and 13 AUC pts. Qualitatively, it outputs non-redundant extractions through the use of its iterative memory (see Table~\ref{tab:redundancy-example}), and a variable number of extractions owing to the \textit{EndofExtractions} token.
It also outperforms CopyAttention with BERT, which is a very strong baseline, by 1.9 Opt. F1 pts, 0.5 AUC and 3.7 Last F1 pts. \shortname{} consistently outperforms CopyAttention with BERT over different bootstrapping datasets (see Table~\ref{tab:bootstrapping-datasets}).

\begin{figure}[t]
\vspace*{-3ex}
\includegraphics[scale=0.5]{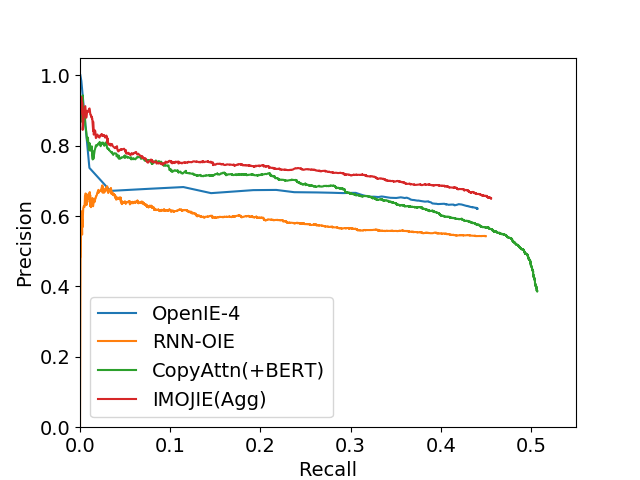}
\caption{Precision-Recall curve of OpenIE Systems.}
\label{fig:pr-curve}
\end{figure}

Figure~\ref{fig:pr-curve} shows that the precision-recall curve of \shortname{} is consistently above that of existing OpenIE systems, emphasizing that \shortname{} is consistently better than them across the different confidence thresholds. We do find that CopyAttention+BERT outputs slightly higher recall at a significant loss of precision (due to its beam search with constant size), which gives it some benefit in the overall AUC. CaRB evaluation of SpanOIE\footnote{\href{https://github.com/zhanjunlang/Span\_OIE}{https://github.com/zhanjunlang/Span\_OIE}} results in (precision, recall, F1) of (58.9, 40.3, 47.9). SpanOIE sources its training data only from OpenIE-4. In order to be fair, we compare it against \shortname{} trained only on data from OpenIE-4 which evaluates to (60.4, 46.3, 52.4). Hence, \shortname{} outperforms SpanOIE, both in precision and recall.

Attention is typically used to make the model focus on words which are considered important for the task. But the \shortname{} model successfully uses attention to \emph{forget} certain words, those which are already covered. 
Consider, the sentence ``He served as the first prime minister of Australia and became a founding justice of the High Court of Australia''. 
Given the previous extraction (He; served; as the first prime minister of Australia), the BERT’s attention layers figure out that the words `prime' and `minister' have already been covered, and thus push the decoder to prioritize `founding' and `justice'.
Appendix~\ref{sec:visualize_attention} analyzes the attention patterns of the model when generating the intermediate extraction in the above example and shows that \shortname{} gives less attention to already covered words.

\vspace{0.5ex}
\subsection{Redundancy}
\emph{What is the extent of redundancy in \shortname{} when compared to earlier OpenIE systems?}

We also investigate other approaches to reduce redundancy in CopyAttention, such as Logician's coverage attention (with both an LSTM and a BERT encoder) as well as diverse beam search. 
Table~\ref{tab:redundancy_rem_exp} reports that both these approaches indeed make significant improvements on top of CopyAttention scores. 
In particular, qualitative analysis of diverse beam search output reveals that the model gives out different words in different tuples in an effort to be diverse, without considering their correctness. 
Moreover, since this model uses beam search, it still outputs a fixed number of tuples.

\begin{table}[h]
\centering {\footnotesize
\begin{tabular}{lccc}
\hline
 \textbf{Extractions} & \multicolumn{3}{c}{\textbf{Metric}}  \\
 & \multicolumn{1}{c}{MNO} & \multicolumn{1}{c}{IOU} & \multicolumn{1}{c}{\#Tuples}\\
\hline
CopyAttention+BERT &  2.805 & 0.463 & 3159 \\ 
\textbf{IMOJIE} & \textbf{1.282} & \textbf{0.208} & \textbf{1620} \\ 
Gold & 1.927 & 0.31 & 2650 \\ \hline
\end{tabular} }
\caption{Measuring redundancy of extractions. MNO stands for Mean Number of Occurrences. IOU stands for Intersection over Union.}
\label{tab:RED}
\end{table}

\begin{table*}
\centering {\footnotesize
\begin{tabular}{lcccc}
\hline
 \textbf{System} & \multicolumn{4}{c}{\textbf{Bootstrapping System}}  \\
 & \multicolumn{1}{c}{OpenIE-4} & \multicolumn{1}{c}{OpenIE-5} & \multicolumn{1}{c}{ClausIE} & \multicolumn{1}{c}{RnnOIE}\\
\hline
Base & 50.7, 29, 50.7 & 47.4, 25.1, 47.4 & 45.1, 22.4, 45.1 & 49.2, 26.5, 49.2 \\ 
CopyAttention+BERT & 51.6, 32.8, 49.6 & 48.7, \textbf{29.4}, 48.0 & 47.4, 30.2, 43.6 & 47.9, 30.6, 41.1 \\ 
\shortname & \textbf{53.2}, \textbf{33.1}, \textbf{52.4} & \textbf{48.8}, 27.9, \textbf{48.7} & \textbf{49.2}, \textbf{31.4}, \textbf{45.5} & \textbf{51.3}, \textbf{31.1}, \textbf{50.8} \\ \hline
\end{tabular} }
\caption{Evaluating models trained with different bootstrapping systems.}
\label{tab:bootstrapping-datasets}
\end{table*}

This analysis naturally suggested the \shortname~(w/o BERT) model --- an \shortname{} variation that uses an LSTM encoder instead of BERT. Unfortunately, \shortname~(w/o BERT) is behind the CopyAttention baseline by 12.1 pts in AUC and 4.4 pts in Last~F1. We hypothesize that this is because the LSTM encoder is unable to learn how to capture \textit{inter-fact dependencies} adequately --- the input sequences are too long for effectively training LSTMs.


This explains our use of Transformers (BERT) instead of the LSTM encoder to obtain the final form of \shortname.  With a better encoder, \shortname{} is able to perform up to its potential, giving an improvement of \textbf{(17.8, 12.7, 19.6)} pts in (Optimal~F1, AUC, Last~F1) over existing seq2seq OpenIE systems.

We further measure two quantifiable metrics of redundancy:
\begin{description}
\item[Mean Number of Occurrences (MNO):] The average number of tuples, every output word appears in.
\item[Intersection Over Union (IOU):] Cardinality of intersection over cardinality of union of words in the two tuples, averaged over all pairs of tuples.
\end{description}

These measures were calculated after removing stop words from tuples.  Higher value of these measures suggest higher redundancy among the extractions.  \shortname{} is significantly better than CopyAttention+BERT, the strongest baseline, on both these measures (Table~\ref{tab:RED}). Interestingly, \shortname{} has a lower redundancy than even the gold triples; this is due to imperfect recall.

\vspace{0.5ex}
\subsection{The Value of Iterative Memory}
\emph{To what extent does the \shortname~style of generating tuples improve performance, over and above the use of BERT?} 

We add BERT to CopyAttention model to generate another baseline for a fair comparison against the \shortname{} model. When trained only on OpenIE-4, \shortname{} continues to outperform CopyAttention+BERT baseline by (1.6, 0.3, 2.8) pts in (Optimal~F1, AUC, Last~F1), which provides strong evidence that the improvements are not solely by virtue of using a better encoder.
We repeat this experiment over different (single) bootstrapping datasets. Table \ref{tab:bootstrapping-datasets} depicts that \shortname~consistently outperforms CopyAttention+BERT model.

We also note that the order in which the extractions are presented to the model (during training) is indeed important. On training IMoJIE using a randomized-order of extractions, we find a decrease of 1.6 pts in AUC (averaged over 3 runs).

\vspace{0.5ex}
\subsection{The value of Score-and-Filter}
\emph{To what extent does the scoring and filtering approach lead to improvement in performance?} 

\shortname{} aggregates extractions from multiple systems through the scoring and filtering approach. It uses extractions from OpenIE-4 (190K), ClausIE (202K) and RnnOIE (230K) to generate a set of 215K tuples. Table \ref{tab:dataset-aggregation1} reports that \shortname{} does not perform well when this aggregation mechanism is turned off. We also try two supervised approaches to aggregation, by utilizing the gold extractions from CaRB's dev set.
\begin{itemize}
    \item \textbf{Extraction Filtering}: For every sentence-tuple pair, we use a binary classifier that decides whether or not to consider that extraction. The input features of the classifier are the \textit{[CLS]}-embeddings generated from BERT after processing the concatenated sentence and extraction. The classifier is trained over tuples from CaRB's dev set.
    \item \textbf{Sentence Filtering}: We use an \shortname{} model (bootstrapped over OpenIE-4), to score all the tuples. Then, a Multilayer Perceptron (MLP) predicts a confidence threshold to perform the filtering. Only extractions with scores greater than this threshold will be considered. The input features of the MLP include the length of sentence, \shortname~ (OpenIE-4) scores, and GPT \cite{Radford2018-GPT} scores of each extraction. This MLP is trained over sentences from CaRB's dev set and the gold optimal confidence threshold calculated by CaRB. 
\end{itemize}
We observe that the Extraction, Sentence Filtering are better than no filtering by by 7.5, 11.2 pts in Last F1, but worse at Opt. F1 and AUC. We hypothesise that this is because the training data for the MLP (640 sentences in CaRB's dev set), is not sufficient and the features given to it are not sufficiently discriminative.
Thereby, we see the value of our unsupervised Score-and-Filter that improves the performance of \shortname{} by (3.8, 15.9) pts in (Optimal F1, Last F1). The 1.2 pt decrease in AUC is due to the fact that the \shortname{} (no filtering) produces many low-precision extractions, that inflates the AUC.

Table \ref{tab:dataset-ablation} suggests that the model trained on all three aggregated datasets perform better than models trained on any of the single/doubly-aggregated datasets. Directly applying the Score-and-Filter method on the test-extractions of RnnOIE+OpenIE-4+ClausIE gives (Optimal F1, AUC, Last F1) of (50.1, 32.4, 49.8). This shows that training the model on the aggregated dataset is important.

\vspace{0.5ex}
\noindent\textbf{Computational Cost}:
The training times for CopyAttention+BERT, \shortname~ (OpenIE-4) and \shortname~ (including the time taken for Score-and-Filter) are 5 hrs, 13 hrs and 30 hrs respectively. This shows that the performance improvements come with an increased computational cost, and we leave it to future work to improve the computational efficiency of these models.


\section{Error Analysis}
We randomly selected 50 sentences from the CaRB validation set. We consider only sentences where at least one of its extractions shows the error. We identified four major phenomena contributing to errors in the \shortname{} model: \newline
(1) \textbf{Missing information:} 66\% of the sentences have at least one of the relations or arguments or both missing in predicted extractions, which are present in gold extractions. This leads to incomplete information. \newline
(2) \textbf{Incorrect demarcation:} Extractions in 60\% of the sentences have the separator between relation and argument identified at the wrong place. \newline
(3) \textbf{Missing conjunction splitting:} In 32\% of the sentences, our system fails to separate out extractions by splitting a conjunction.  E.g., in the sentence ``US 258 and NC 122 parallel the river north \dots'', \shortname{} predicts just one extraction (US 258 and NC 122; parallel; \dots) as opposed to two separate extractions (US 258; parallel; \dots) and (NC 122; parallel; \dots) as in gold. \newline
(4) \textbf{Grammatically incorrect extractions:} 38\% sentences have a grammatically incorrect extraction (when serialized into a sentence).  \newline
Additionally, we observe 12\% sentences still suffering from \textbf{redundant} extractions and 4\% \textbf{miscellaneous} errors.

\section{Conclusions and Discussion}
\label{sec:End}

We propose \shortname~for the task of OpenIE. \shortname~significantly improves upon the existing OpenIE systems in all three metrics, Optimal F1, AUC, and Last F1, establishing a new State Of the Art system. Unlike existing neural OpenIE systems, \shortname~produces non-redundant as well as a variable number of OpenIE tuples depending on the sentence, by iteratively generating them conditioned on the previous tuples. Additionally, we also contribute a novel technique to combine multiple OpenIE datasets to create a high-quality dataset in a completely unsupervised manner. We release the training data, code, and the pretrained models.\footnote{\href{https://github.com/dair-iitd/imojie}{https://github.com/dair-iitd/imojie}}

\shortname~ presents a novel way of using attention for text generation. \citet{bahdanau&al15} showed that attending over the input words is important for text generation. \citet{see&al17} showed that using a coverage loss to track the attention over the decoded words improves the quality of the generated output. We add to this narrative by showing that deep inter-attention between the input and the partially-decoded words (achieved by adding previous output in the input) creates a better representation for iterative generation of triples. This general observation may be of independent interest beyond OpenIE, such as in text summarization.

{\footnotesize
\section*{Acknowledgements}
 IIT Delhi authors are supported by IBM AI Horizons Network grant, an IBM SUR award, grants by Google, Bloomberg and 1MG, and a Visvesvaraya faculty award by Govt. of India. We thank IIT Delhi HPC facility for compute resources. Soumen is supported by grants from IBM and Amazon. We would like to thank Arpita Roy for sharing the extractions of SenseOIE with us.
}


\bibliography{anthology,acl2020}
\bibliographystyle{acl_natbib}

\clearpage

\appendix

\section*{\intitle\\ (Supplementary Material)}

\section{Performance with varying sentence lengths}
\begin{figure}[h]
\begin{center}
\advance\leftskip-3cm
\advance\rightskip-3cm
\includegraphics[keepaspectratio=true,scale=0.4]{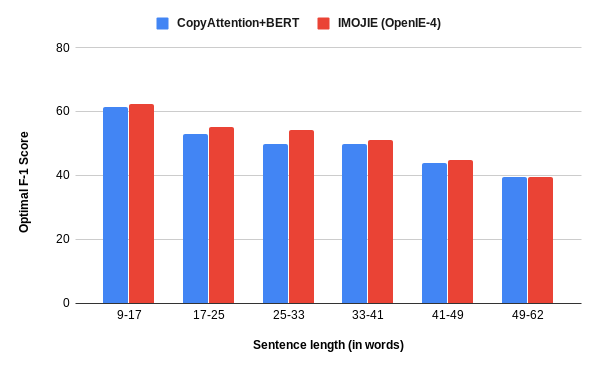}
\caption{Measuring performance with varying input sentence lengths}
\label{fig1}
\end{center}
\end{figure}

In this experiment, we measure the performance of baseline and our models by testing on sentences of varying lengths. We partition the original CaRB test data into 6 datasets with sentences of lengths (9-16 words), (17-24 words), (25-32 words), (33-40 words), (41-48 words) and (49-62 words) respectively. Note that the minimum and maximum sentence lengths are 9 and 62 respectively. We measure the Optimal F1 score of both Copy Attention + BERT and IMOJIE (Bootstrapped on OpenIE-4) on these partitions as depicted in Figure \ref{fig1}. \newline
We observe that the performance deteriorates with increasing sentence length which is expected as well. Also, for each of the partitions, IMOJIE marginally performs better as compared to Copy Attention + BERT.

\section{Measuring Performance on Varying Beam Size}
\begin{figure}
\begin{center}
\advance\leftskip-3cm
\advance\rightskip-3cm
\includegraphics[keepaspectratio=true,width=\hsize]{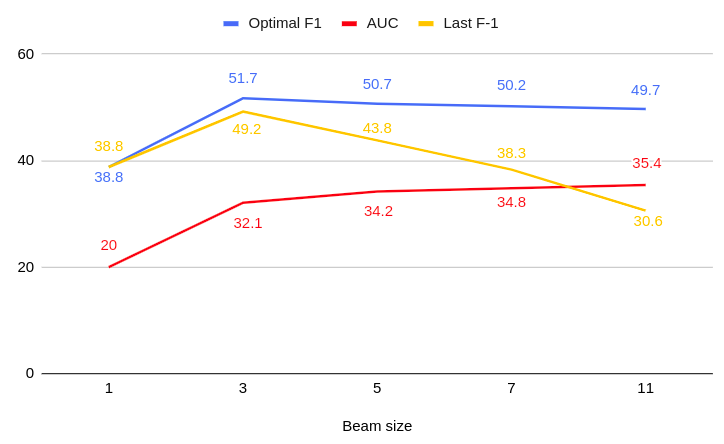}
\caption{Measuring performance of CopyAttention with BERT model upon changing the beam size}
\label{fig2}
\end{center}
\end{figure}

We perform inference of the CopyAttention with BERT model on CaRB test set with beam sizes of 1, 3, 5, 7, and 11. We observe in Figure \ref{fig2} that AUC increases with increasing beam size. A system can surge its AUC by adding several low confidence tuples to its predicted set of tuples. This adds low precision - high recall points to the Precision-Recall curve of the system leading to higher AUC.\\
On the other hand, Last F1 experiences a drop at very high beam sizes, thereby capturing the decline in performance. Optimal F1 saturates at high beam sizes since its calculation ignores the extractions below the optimal confidence threshold.\\
This analysis also shows the importance of using Last F1 as a metric for measuring the performance of OpenIE systems.

\begin{table*}
\begin{center} {\footnotesize
\begin{tabular}{lccc}
\hline
 \textbf{Model} & \multicolumn{3}{c}{\textbf{Dataset}}  \\
 & \multicolumn{1}{c}{Wire57} & \multicolumn{1}{c}{Penn} & \multicolumn{1}{c}{Web}\\
\hline 
CopyAttention + BERT & 45.60, \textbf{27.70}, 39.70 & 18.20, 7.9, 12.40 & 30.10, \textbf{18.00}, 14.60 \\ 
\shortname & \textbf{46.20}, 26.60, \textbf{46.20} & \textbf{20.20}, \textbf{8.70}, \textbf{15.50} & \textbf{30.40}, 15.50, \textbf{26.40} \\ \hline
\end{tabular} }
\end{center}
\caption{Evaluation on other datasets with the CaRB evaluation strategy}
\label{tab:ODE}
\end{table*}

\section{Evaluation on other datasets}

We use sentences from other benchmarks with the CaRB evaluation policy and we find similar improvements, as shown in Table \ref{tab:ODE}. \shortname{} consistently outperforms our strongest baseline, CopyAttention with BERT, over different test sets. This confirms that \shortname{} is domain agnostic.

\section{Visualizing Attention}
\label{sec:visualize_attention}

Attention has been used in a wide variety of settings to help the model learn to focus on important things \cite{bahdanau&al15, xu&al15, lu&al19}. However, the \shortname{} model is able to use attention to understand which words have already been generated, to focus on remaining words. In order to understand how the model achieves this, we visualize the learnt attention weights. There are two attention weights of importance, the learnt attention inside the BERT encoder and the attention between the decoder and encoder.  We use BertViz \cite{vig@19} to visualize the attention inside BERT.

We consider the following sentence as the running example - "he served as the first prime minister of australia and became a founding justice of the high court of australia". We visualize the attention after producing the first extraction - ``he; served; as the first prime minister of australia''. Intuitively, we understand that the model must focus on the words ``founding'' and ``justice'' in order to generate the next extraction - ``he; became; a founding justice of the high court of australia''. In Figure \ref{fig:prime} and Figure \ref{fig:minister} (where the left-hand column contains the words which are used to attend while right-hand column contains the words which are attended over), we see that the words ``prime'' and ``minister'' of the original sentence have high attention over the same words in the first extraction. But the attention for ``founding'' and ``justice'' are limited to the original sentence.

Based on these patterns, the decoder is able to give a high attention to the words ``founding'' and ``justice'' (as shown in Figure \ref{fig:att_weights}), in-order to successfully generate the second extraction "he; became; a founding justice of the high court of australia".

\begin{figure}
\begin{center}
\advance\leftskip-3cm
\advance\rightskip-3cm
\includegraphics[keepaspectratio=true,width=\hsize]{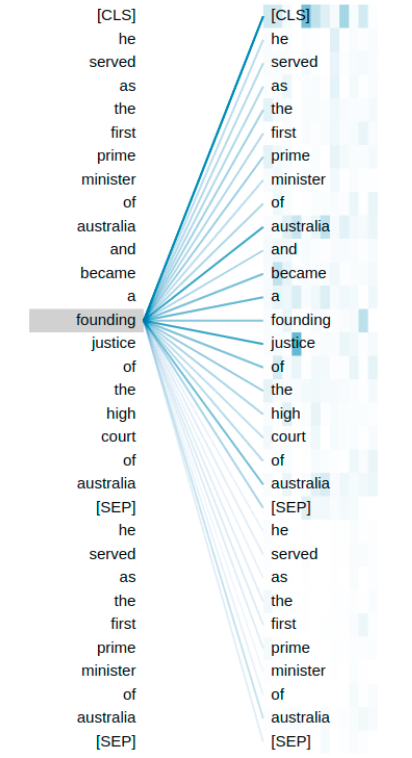}
\caption{BERT attention for the word `founding'}
\label{fig:founding}
\end{center}
\end{figure}

\begin{figure}
\begin{center}
\advance\leftskip-3cm
\advance\rightskip-3cm
\includegraphics[keepaspectratio=true,width=\hsize]{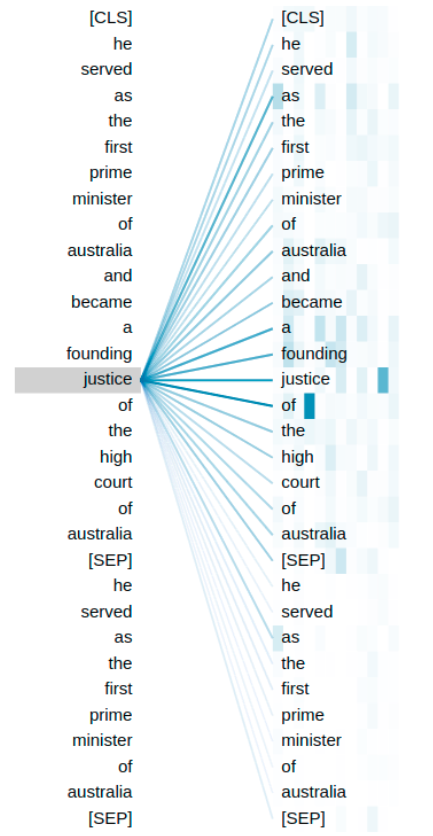}
\caption{BERT attention for the word `justice'}
\label{fig:justice}
\end{center}
\end{figure}

\begin{figure}
\begin{center}
\advance\leftskip-3cm
\advance\rightskip-3cm
\includegraphics[keepaspectratio=true,width=\hsize]{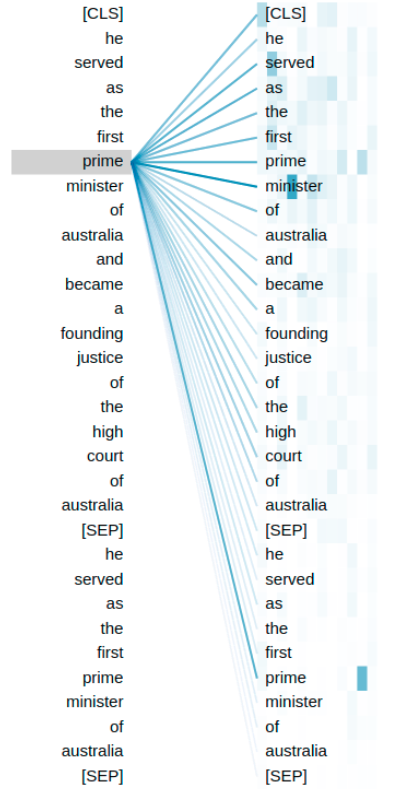}
\caption{BERT attention for the word `prime'}
\label{fig:prime}
\end{center}
\end{figure}

\begin{figure}
\begin{center}
\advance\leftskip-3cm
\advance\rightskip-3cm
\includegraphics[keepaspectratio=true,width=\hsize,scale=1.5]{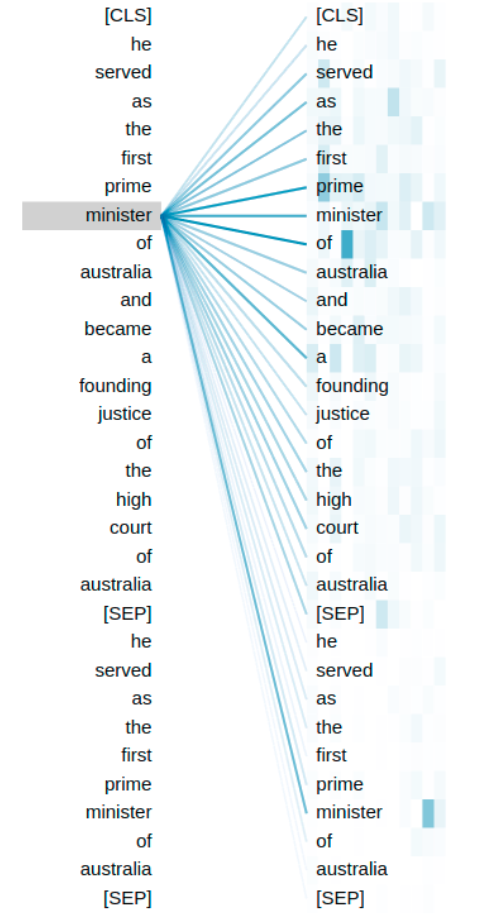}
\caption{BERT attention for the word `minister'}
\label{fig:minister}
\end{center}
\end{figure}

\begin{figure*}
\begin{center}
\advance\leftskip-3cm
\advance\rightskip-3cm
\includegraphics[keepaspectratio=true,width=\hsize]{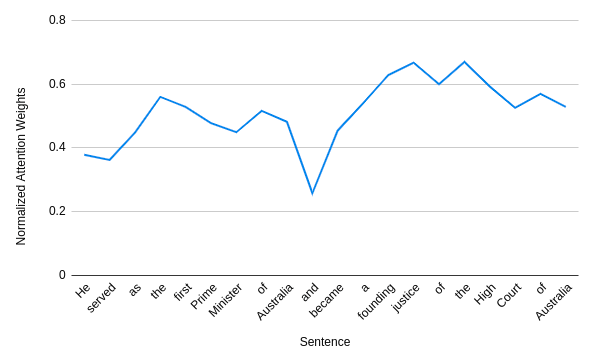}
\caption{Attention weights for the decoder}
\label{fig:att_weights}
\end{center}
\end{figure*}

\end{document}